\documentclass{ws-procs11x85}
\usepackage{ws-procs-thm}  
\usepackage[caption=false]{subfig}
\begin{document}

\title{ParKCa: Causal Inference with Partially Known Causes}

\author{Raquel Aoki$^\dag$ and Martin Ester}

\address{School of Computing Science, Simon Fraser University\\
  Burnaby, Canada\\
$^\dag$E-mail: raoki@sfu.ca}

\begin{abstract}
Methods for causal inference from observational data are an alternative for scenarios where collecting counterfactual data or realizing a randomized experiment is not possible. Our proposed method ParKCA combines the results of several causal inference methods to learn new causes in applications with some known causes and many potential causes. We validate ParKCA in two Genome-wide association studies, one real-world and one simulated dataset. Our results show that ParKCA can infer more causes than existing methods.
\end{abstract}


\copyrightinfo{\copyright\ 2016 The Authors. Open Access chapter published by World Scientific Publishing Company and distributed under the terms of the Creative Commons Attribution Non-Commercial (CC BY-NC) 4.0 License.}

\section{Introduction}\label{introduction}

The vision of precision medicine is the development of prevention and treatment strategies that take individual variability into account  \cite{collins2015new}. Precision medicine promises to allow more precise diagnosis, prognosis, and treatment of patients, based on their individual data. In the context of precision medicine, it is important to understand the leading causes of the outcome of interest, which can be achieved using causal discovery methods. Drug response and adversarial drug reactions are examples of an outcome of interest, and OMICS data record potential causes and confounders.

The gold standard of causal inference is based on experimental design, with randomized trials and control groups, which is not always available due to the lack of the full experiment and counterfactual data, either because it is too expensive or impossible to collect. Therefore, there is a rise of more data-driven methods, based on observational data, to either perform causal discovery or estimate treatment effects\cite{johansson2016learning,louizos2017causal,wang2019blessings}.  Furthermore, some applications, such as Driver Gene Discovery\cite{stratton2009cancer}, have a few causes that are well known. Most of the existing methods use these only to evaluate or to eliminate edges on constraint-based causal discovery methods. 

To handle computational biology (CB) applications with thousands of treatments, unobserved confounders, and partially known causes, we propose ParKCa: a method that uses the few known causes to learn new causes through the combination of causal discovery methods. The intuition is that ParKCa will learn how to identify causes based on the outputs of the other methods (similar to ensemble learning) and a few known examples. ParKCa has many advantages. First, leveraging several methods instead of using a single one can minimize biases and highlight patterns common across the methods. Second, it allows the use of known causes to help identify new causes. Finally, it also allows the combinations of several datasets that share the same set of possible causes but might differ in the datatype or set of rows.

The proposed method ParKCa is validated in a simulated dataset and on the driver gene discovery application. The existence of associations between cancer and specific genes is well accepted in the precision medicine field. However, the human body has more than 20,000 genes, and not all genes mutations lead to cancer\cite{stratton2009cancer}. Hence, the challenge here is to recognize those genes that are associated with cancer spreading from the original site to other areas of the body (metastasis) through causal inference. These genes are known as \textit{driver genes} and play an important role in cancer prevention and treatment.
    
There are many challenges around driver gene discovery. The progress in sequencing technology and the lower cost of collecting genetic information allowed the creation of datasets such as The Cancer Genome Atlas (TCGA).  However, the number of columns (genes) is often much larger than the number of rows (patients), which poses a challenge for machine learning models. The partially known dependence between genes due to pathways is also a challenge. Pathways are sets of genes where the alteration or mutation in one gene can cause changes in other genes that share the same pathway\cite{vandin2012novo}. Additionally, some elements that cause cancer might not be included in the dataset. Examples of attributes not observed are the structured clinical information about the patient, such as their lab results and lifestyle. Finally, the lack of a well-defined training set is also a challenge that makes the evaluation of results tricky. There is no `true' list of driver genes to evaluate the quality of the machine learning models\cite{dees2012music,schroeder2014oncodriverole,tokheim2016evaluating}. 

\begin{table}
\tbl{Toy example of the transposed input data (on the left) and output data (on the right). Note that in the input data we have $Y$ and in the outcome data, the \textit{known causes}.}
{\begin{tabular}{l|ccc|cccl|cc|c}
\cline{0-4}\cline{8-11}
& Gene 1 & ... & Gene V & Y &  & & & $L_1$ &  $L_2$ & Known Cause \\\cline{0-4}\cline{8-11}
 Patient 1 & 7.39 & ... & 1.60 & 0  &  & & Gene 1    & -1.2 &  -2.4 & 1 \\
 ...       & ... & ... & ... & ...  & $\rightarrow$ & & ...       & ...  & ... & ..\\
 Patient J & 3.25 & ... & 2.73 & 1  & & & Gene V    & 0 &  12.3 & 0\\\cline{0-4}\cline{8-11}
\end{tabular}}
\label{inputdataset}
\end{table}

A toy example is shown in Table \ref{inputdataset}. The goal is to learn which genes among $V$ genes are causally associated with a phenotype $Y$ from a dataset with $J$ patients as in the left side of Table \ref{inputdataset}. The input data represents the gene expression of patients, and it is used to fit the level 0 models $L_1$ and $L_2$. The right side of Table \ref{inputdataset} shows the output data, constructed with the learners' output. We add to the output an attribute with the partially known causes. 
    
The main contributions of this paper are as follows: 
\begin{itemize}
    \item We introduce the problem of causal discovery from observational data with partially known causes. We are the first ones to formalize it as a stacking problem.
    \item We propose ParKCa, a flexible method that learns new causes from the outputs of causal discovery methods and from partially known causes.
    \item ParKCa is validated on a real-world TCGA dataset for identifying genes that are potential causes of cancer metastases and on simulated genomic datasets. 
\end{itemize} 
    
\section{Related Work}

Our work combines several research areas: 

\textbf{Causality}: Motivated by the need for models that are more robust, reproducible, and easier to explain, causality has received a lot of attention. Constraint-based and score-based causal discovery methods, such as PC-algorithm\cite{spirtes1991algorithm}, fast PC-algorithm\cite{le2016fast}, FCI\cite{spirtes2000causation}, RFCI\cite{colombo2012learning}, and fGES\cite{ramsey2017million}, are still largely used. Their main goal is to recover the causal structure that fits the observed data. However, these methods have a poor performance on large dimensional datasets and/or assume causal sufficiency, suppositions that fail on most of the computational biology (CB) applications. 

Deep learning models are making significant contributions to the estimation of treatment effects in the past years\cite{johansson2016learning,louizos2017causal,shi2019adapting}. BART\cite{hill2011bayesian} uses the Conditional Average Treatment Effect (CATE) to estimate the treatment effects has been successful in many applications. Finally, the Deconfounder Algorithm (DA)\cite{wang2019blessings}, combines probabilistic factor models and outcome models to estimate causal effects. Considering the challenge of learning causes from several datasets, Tillman and Spirtes\cite{tillman2011learning} proposed a method to learn equivalence classes from multiple datasets. A limitation of this work, however, is the lack of scalability to large dimensional datasets. 

\textbf{Ensembles}: Our work is based on stacking ensemble\cite{wolpert1992stacked}. Its main idea is to use several learners models whose outputs are combined and used as input for fitting a meta-model. The idea of using an ensemble approach to calculate causal effects is not new\cite{schuler2017targeted,kunzel2019metalearners}. 
Instead of using ensemble learning to make more accurate causal effect estimates, our work focuses on using ensemble learning to discover new causes. We adopted commonly used meta-learner models (Logistic Regression, Random Forest, Neural Networks, and others), and we also explore PU-learning classification models\cite{liu2003building,elkan2008learning, du2015convex}, a sub-class of semi-supervised learning. Our method performs a classification task with the meta-learner on the level 1 data $D^1_{V\times L}$ and a new variable that encodes the \textit{known causes}. The labels are $Y_v^{1}=1$ for well-known causes and $Y_v^{1}=0$ for non-causal or unknown causes. In other words, the classification learns from positive (known causes) and unlabeled (not causal or unknown causes) data.

\textbf{Driver Gene Discovery (real-world dataset)}: Spurious correlations or associations between genes and metastasis are common. Therefore, the challenge lies in identifying those genes that are true causes (\textit{driver genes}) of the underlying condition, not just associated with it. In our real-world dataset, we want to find genes that contribute to cancer metastasis development. In this condition, cancer spreads from the original site to other areas. Previous methods that explored this application are: MuSiC\cite{dees2012music}, OncodriveFM\cite{gonzalez2012functional}, ActiveDriver\cite{reimand2013systematic}, TUSON\cite{davoli2013cumulative}, OncodriveCLUST\cite{tamborero2013oncodriveclust}, MutsigCV\cite{lawrence2014discovery}, OncodriveFML\cite{mularoni2016oncodrivefml}, 20/20+ \url{(https://github.com/KarchinLab/2020plus)}, and others\cite{vandin2012novo, schroeder2014oncodriverole}. The first challenge of this application is the large number of genes (possible causes) along with the small sample size and the known (and unknown) dependencies among genes, which adds a certain complexity to the problem. The existence of confounders, some possible to be observed (such as clinical information), others not (such as family history or lifestyle)\cite{stratton2009cancer} poses another challenge. Finally, the limited and biased list of known driver genes\cite{futreal2004census} also needs some attention\cite{tokheim2016evaluating}. This list, here referred to as Cancer Gene Census (CGC), is the gold-standard of driver genes currently available.

\section{The ParKCa Method}\label{method}

\begin{figure}[!h]
    \centering
    \includegraphics[scale=0.55]{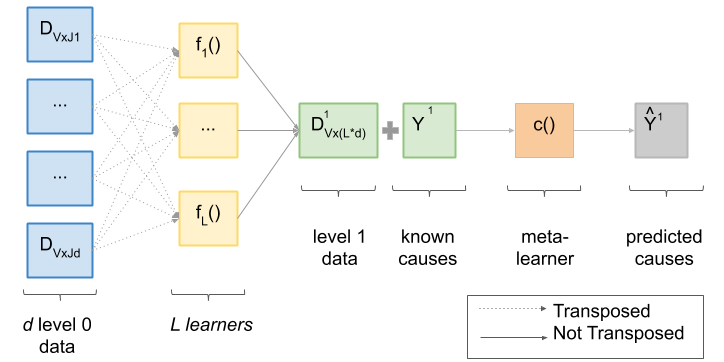}
    \caption{Illustration of the ParKCa method. From $d$ datasets/subsets with $J_d$ columns (examples) and $V$ rows (possible causes), $L\times d$ outputs are extracted using $L$ level 0 models. These outputs are aggregated in a single dataset $D^{1}_{V\times (L*d)}$. In this step, we also add the partially known causes $Y^{1}$ and fit a meta-learner model to predict new potential causes.}
    \label{fig1}
\end{figure}

ParKCa deals with causal discovery from a stacking ensemble perspective with some adaptations. Typically, each causal discovery method is estimated individually, and their results compared. In ParKCa, we use the causal discovery methods' outputs as a classifier's features to learn how these methods agree to identify new causes based on a few known causes used as examples. According to the stacking nomenclature, the causal discovery methods are our learners, and the classification model is our meta-learner, as shown in Figure \ref{fig1}.

Compared to a standard stacking model, the first modification required by our approach is how we feed the learners. Unlike stacking used as a predictive model, where the goal is to maximize the accuracy of the predictions, we focus on the features (potential causes). Therefore, a learners $f_L$ receives a $transpose(D^0_{V\times J})$ as input data. The outcome of interest $Y_{J}^0$ can be easily added using the transposed level 0 data. In our real-world dataset, $Y^0$ encodes if the patient had cancer metastasis or not. The learners'output is one value per feature $v\in\{1,...,V\}$. The second modification is how we create level 1 data. Traditionally, to avoid data leaking and overfitting, stacking learning models use cross-validation to make the predictions used on level 1 data. The same is not possible in our approach: subsets of the possible causes would violate assumptions, such as causal sufficiency. Instead, we use bootstrapping on the transposed level 0 data. By doing so, we can decrease biases and test the significance of coefficients when suitable. 

All assumptions that the learners make, such as discrete treatments, also need to be satisfied. ParKCa requires the number of possible causes $V$ to be sufficiently large to be able to fit the meta-learner. As an ensemble method, ParKCa requires sufficient diversity among the learners. We check the diversity with the averaged Q statistics\cite{kuncheva2003measures} over all pairs of classifiers. Considering two learners, $f_i$ and $f_j$, and the number of True Positives (TP), True Negatives (TN), False Positives (FP) and False Negatives (FN) from a confusion matrix between the two learners, the Q Statistic is $Q_{i,j} = \frac{TP\times TN - FP\times FN}{TP\times TN + FP\times FN}$, and $Q_{i,j}\in[-1,1]$. The diversity increases when the learners commit errors in different objects, resulting in a negative or close to zero $Q_{i,j}$. For L learners, the average $Q$ is defined as:  

 \begin{equation}\label{q}
     Q_{av} = \frac{2}{L(L-1)}\sum_{i=1}^{L-1}\sum_{j=i+1}^LQ_{i,j}
 \end{equation}

\subsection{Learners}\label{Learners}

ParKCa starts by fitting the learners, also called level 0 models. As Figure \ref{fig1} shows, the $transpose(D^0_{V\times J})$ is the input of the learners $f_l, \forall l\in[1, L]$. The level 0 models of ParKCA are causal discovery methods or models that estimate the treatment effect, and their output concatenated is the level 1 data $D^1_{V\times L}$. The learners employed must have all their assumptions satisfied for the validity of the results. The Deconfounder Algorithm (DA)\cite{wang2019blessings}, for example, requires a predictive check of its latent variables. Therefore, it is necessary to verify if the factor model passes the predictive check.

Defining $\phi_{(v,l)}$ as the outcome from learner $f_l$ and potential cause $v$, the value $\phi_{(v,l)}$ can be continuous or discrete depending on the outputs of the learners. The outcomes $\phi_{(v,l)}\forall v \in V$ and $\forall l \in L$ aggregated form the level 1 data $D^1_{V\times L}$. Optionally, one might choose to set all non-causal variables to $0$. In this case, if $v$ is a non-causal variable according to method $f_l$, then  $D^1_{v\times l} = 0$, else, $D^1_{v\times l}=\phi_{(v,l)}$. 

Some methods, such as models that estimate treatment effects, might benefit from using bootstrapping to decrease biases and, when suitable, perform a statistical test for $H_0: \phi_{(v,l)} = 0$ or $H_1: \phi_{(v,l)} \neq 0$. To use bootstrap, we suggest the following steps: 

\begin{enumerate}
        \item Take $B$ samples  of size $J' = \lfloor 0.9*J\rfloor$ from $transpose(D^0_{J\times V})$ and fit the learner $f_l$ in each sample $D^0_{J'\times V}$ saving the estimated outputs $\phi_{(v,l),b}$; then, set $D^1[v,l] = \frac{1}{B}\sum_{b=1}^B{\phi_{(v,l),b}}$ (Strong Law of Large Numbers)
        \item \textit{(Optional)} Apply a two-tailed test to check the hypothesis test with the sample $\{\phi_{(v,l),1},...,\phi_{(v,l),B}\}$. If $p$-value $\leq 0.05$, reject $H_0$ and set $D^1[v,l] = \frac{1}{B}\sum_{b=1}^B{\phi_{(v,l),b}}$, else $0$. 
\end{enumerate}

According to the Strong Law of Large Numbers, let $\{\phi_{(v,l),1},...,\phi_{(v,l),B}\}$ be independent identically distributed random variables with $E|\phi_{(v,l),b}|<\infty$, then the average of the samples converges to the true mean when B is sufficient large.  

ParKCa assumes that the number of possible causes $V$ is sufficiently large to fit a meta-learner. Therefore, the learners must be robust to large datasets. A few examples are the RFCI\cite{colombo2012learning} and fGES\cite{ramsey2017million} work well in applications where there are no unobserved confounders; the PC-algorithm fast\cite{le2016fast} is robust to unobserved confounders, but its performance in large datasets is poor; the DA\cite{wager2018estimation} and CEVAE\cite{louizos2017causal} are suitable to applications with unobserved confounders. 

To validate ParKCa, in the experiments we worked with three methods: the Deconfounder Algorithm (DA)\footnote{There is an ongoing discussion about DA, with some recent criticism 
\cite{ogburn2019comment,d2019comment} and extra clarification\cite{wang2019blessingsreplay} presented. ParKCa assumes that the original work\cite{wang2019blessings} is correct. However, in case the reader is uncomfortable with the use of this method, we recommend to replace it with other suitable learner.}\cite{wang2019blessings}, BART\cite{hill2011bayesian}, and CEVAE\cite{louizos2017causal}. The main idea behind DA is to learn latent features as a substitute for unobserved confounders. Then, use the data augmented with the latent variables to make the causal inference through an outcome model. The use of proxies to replace true confounders in causal inference analysis\cite{kallus2018causal} will also be employed on the BART model. BART makes data interventions to estimate the conditional average treatment effect (CATE). For each possible cause $v\in\{1,...,V\}$:  
\begin{equation}
CATE_v = E[Y|X=x,do(X_v=a),Z] - E[Y|X=x,do(X_v=0),Z]
\end{equation}
where $X_v=0$ represents the intervention component on the observed data  $X_v=a$ and $Z$ the estimated proxies. Finally, CEVAE infers causal effects from observational data and is robust to unobserved confounders. Based on Variational Autoencoders (VAE), it tries to simultaneously discover the hidden confounders and infer how they affect the treatment and output. 

The learners of our experiments were selected to satisfy the requirements and assumptions of the application. We would like to emphasize that ParKCa is not limited to these learners.

\subsection{Meta-learner}\label{metaleaner}

The level 1 dataset $D^1_{V\times L}$ records the outputs of $L$ learners for $V$ possible causes. If multiple level 0 datasets are being used, then the format is $D^1_{V\times L*d}$, where $d$ is the number of level 0 datasets (see Figure \ref{fig1}). The prior knowledge about known causes is added as a new attribute $Y^1$, where $Y^1_v = 1$ if $v$ is a known cause, and 0 otherwise. Note that, unless all the possible causes are known, some true causes will be labeled as $0$. ParKCa uses binary classification models as meta-learners. The level 1 data contains only positive or unlabeled examples, so we tested PU-learning classification models and compared their results with traditional classification models (Logistic Regression (LR), Random Forest (RF), and Neural Network (NN)). 

The PU-learning model Adapter-PU\cite{elkan2008learning}  uses a traditional probabilistic classifier $c^o(X)$ such that $c^o(X)=p(Y = 1|X)$ is as close as possible. This method assumes that the labeled positive examples are randomly selected among all true positive examples. Unbiased PU (UPU)\cite{du2015convex} is another PU-learning model adopted. UPU is a convex classification method that aims to cancel the bias from the unlabeled data being a mix of positive and negative examples by using a loss function for positive examples and another loss function for unlabeled examples.

The traditional binary classification models consider all unlabeled examples as negatives examples or non-causal variables, which can add bias and noise to the predictions. A majority vote ensemble from the methods described above is also used. Finally, a random model is also compared. The random model assigns the labels 1 and 0 according to the proportion of 1's and 0's in the training data.

\section{Experiments}
We performed experiments to validate our method on two Genome-wide association studies (GWAS) datasets, a real-world dataset, and a simulated dataset. \\
\textbf{Real-world dataset:} We use The Cancer Genome Atlas Program (TCGA) dataset, which has available the gene expression (RNA-seq) of patients with cancer. The data pre-processing is described in the Supplemental Material A.1, and the level 0 dataset from this application has 7066 genes and 2854 patients, of which 1039 (36\%) have metastases. From the 7066 genes, 681 (9\%) are known driver genes\cite{futreal2004census}. These known driver genes are our positive examples in the meta-learner models. For this application, we also worked with multiple datasets considering their clinical information, such as gender and cancer type. \\
\textbf{Simulated datasets:} We simulated GWAS data following the scheme described by  Wang and
Blei\cite{wang2019blessings} and Song et al.\cite{song2015testing}, illustrated with more details in the  Supplemental Material A.2. Single nucleotide polymorphisms (SNPs), the most common type of genetic variation among people, is the datatype adopted. We simulated 10 independent datasets, with 5000 individuals and 10000 SNPs, and confounders. 10\% of these SNPs were set to be causal of a binary trait. 

To validate our method, we first evaluate the learners adopted. Then, we check if ParKCa indeed contributes to detecting more causes. We also verify for the simulated dataset if ParKCa can be used to make better estimates of the treatment effect. Finally, as an extra analysis, we compare our results in the real-world dataset with the state-of-art methods in driver gene discovery. We say `extra' because these methods are not standard causal discovery methods but aim to solve the same problem.

\textbf{Evaluation of the learners}: We adopted DA with the probabilistic PCA\cite{tipping1999probabilistic} as a factor model and Logistic Regression with the elastic net as an outcome model. In our experiments, increasing the number of the latent variables did not improve the results (See Supplemental Material B.1); thus, we adopted $k=15$. The DA models passed the predictive check with $k=15$ (average $p$-value$=0.7181$ at real-world dataset and $0.5381$ on the simulated dataset). 

The ROC curves of the learners on level 0 data were also evaluated (Supplemental Material B.2). To construct the ROC curves, we split the transposed dataset $D_{V\times J}^0$ into training (67\%) and testing set (33\%). Thus, each set has all the possible causes and a subset of the samples. Using the models fitted on the training set, we predicted the outputs for the testing set, which is the outcome of interest at level 0. We used 12 learners on the real-world dataset: BART $+$ 3 datasets (all patients, female and male patients), and DA $+$ 9 datasets (all patients, female and male patients, and six groups of patients with same cancer type). All level 0 models had an excellent performance, except for 4 DA models constructed using datasets based on cancer type. Therefore, we removed the outputs obtained through datasets based on cancer types ESCA, LIHC, PAAD, and SARC from the level 1 dataset. In the simulation study, we repeated the experiment ten times, once for each simulated dataset, and reported the rate of True Positives and False Positives using either CEVAE or DA for each repetition. All learners had a good performance in predicting the outcome of the level 0 datasets. The CEVAE's convergence plots indicate that the model converges after 40 epochs on average. Some randomly selected convergence plots are shown in the Supplemental Material B.3. 

The last step of the learners' evaluation is checking the diversity among the final level 0 models measured by $Q_{av}$. Negative values of  $Q_{av}$ indicate diversity because the learners are committing errors on different objects. The learners used on the real-world dataset about driver gene discovery has diversity equal to $-0.013$, and the average diversity of the learners used with the simulated dataset was $-0.051$, from which we conclude there is diversity among the learners adopted in our experiments. 

\textbf{Learners \textit{versus} Meta-Learners}: This comparison is the core of our validation process. Here, we investigate if adding an extra layer in ParCKa, the meta-learner, contributes to discovering more causes. Therefore, we compare the learners and the meta-learners capacity of identifying causal variables. The learners are used individually to predict if a variable is causal according to their metrics and definitions. These predictions are compared with the meta-learners' predictions $\hat{Y}^1$.

\begin{figure}[!h]
    \subfloat[\textit{Real-world dataset:} Comparison between learners and ParKCa meta-learners to recover causes. ]{\label{app_2}\includegraphics[width=0.5\textwidth]{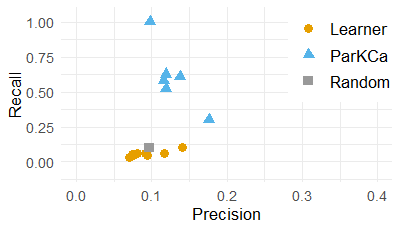}}
    \subfloat[\textit{Simulated dataset:} Comparison to identify causes with different proportions of known causes. Large F1-score indicate good models.]{\label{sim_c}\includegraphics[width=0.5\textwidth]{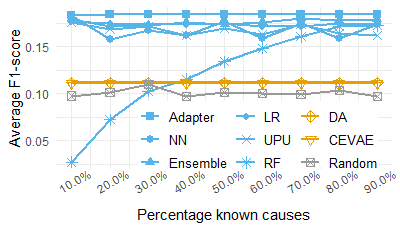}}
    \quad
    \subfloat[\textit{Simulated dataset:} PEHE on the level 1 test set. Small PEHE indicates good models.]{\label{sim_5}\includegraphics[width=0.96\textwidth]{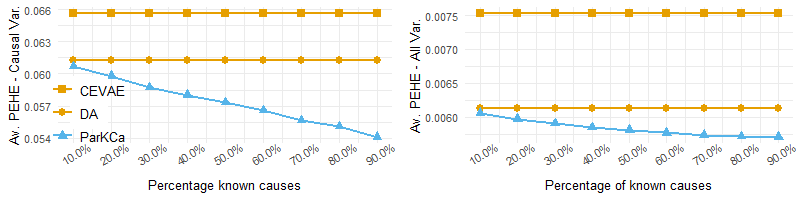}}

    \caption{Causal discovery task evaluation. The learners consider only the original data; ParKCA meta-learners use the learners' outputs and partially known causes to identify more causes.}
    \label{exp1}
\end{figure}

The DA considers variables significantly different from 0 in the outcome model as a causal variable. BART and CEVAE do not have a similar metric, and they only provide treatment effect estimates. One option is using the bootstrap method explained earlier to fit a confidence interval and check if the treatment effect estimated is significantly different from zero. In our experiments, however, we obtained better performance by assigning the 10\% largest estimated treatment effects as causal variables and setting the other variable as non-causal. The learners are compared individually against 6 meta-learners (UPU, Adapted-PU, Logistic Regression - LR, Random Forest - RF, Neural Network - NN, Ensemble - E), and a random model. 

To perform a fair comparison between learners and meta-learners, we split the level 1 data into training and testing sets. We calculated the precision and recall using only the predicted values for the testing set, and the list of known causes as ground truth. The results for the real-world data are shown in Figure \ref{app_2}. While meta-learners and learners models have similar precision, meta-learners tend to have much better recall then learners. Overall, ParKCa has fewer causal variables undetected (False Negatives) than the learners. Figure \ref{sim_c} shows the average F1-score on the simulated datasets versus the proportion of known causes. The proportion of known causes represents how much of the true causes ParCKa has access: if ParKCa knows 40\% of the causes, during the training phase, we randomly label 40\% of the true causes as 1 and the other 60\% as 0, to replicate what we usually encounter in the real-world.We observe that even when ParKCa can access to only a small proportion of true causes, it performs better than the existing baselines, which are independent of the percentage of known causes. The meta-learners, except for the RF, have a higher average F1-score even when only 10\% of the causes are known. These results validate our claim that the stacking approach used by ParKCa can identify more causes then existing methods when some causes are known. All meta-learners, except the RF, seem to be independent of the proportion of known causes on the testing set, which points out another ParKCa's quality: even a small portion of known causes can produce better results than traditional methods. These results sustain our claim that the stacking approach (ParKCa) is capable of identifying more causes than isolated causal methods (learners).  

\textbf{Treatment Effect Estimates}: We investigate a secondary result of the learners used on the simulated datasets, the estimation of treatment effect. We compare the Precision in Estimation of Heterogeneous Effect (PEHE)\cite{hill2011bayesian, louizos2017causal} of our approach against that of the learners used in dependency from the percentage of known causes. $PEHE = \frac{1}{N}\sum_{i=1}^N((y_{i1}-y_{i0})-(\hat{y}_{i1}-\hat{y}_{i0}))$, where $y_{i1}$ and $y_{i0}$ are the true treatment effects, and $\hat{y}_{i1}$ and $\hat{y}_{i0}$ are the estimated treatment effects. This scenario requires known treatment effect estimates, which are hardly ever available in real-world applications. However, this experiment can easily be performed on a simulated dataset, and its results collaborate with our claim that ParKCa finds better results than the learners. We adopted a simple linear regression model as a meta-learner, and we split the level 1 data into training and testing sets. We compared the PEHE of the meta-learner with CEVAE and DA on the test set. Figure \ref{sim_5} shows the PEHE for the causal variables in the left and for all variables on the right. The average PEHE for the ParKCa meta-learner is similar to that of DA when only 10\% of the causes are known; however, it decreases when more causes are known in both plots. These results point out an alternative use of ParKCa for treatment effect estimation.

\begin{figure}[!h]
    \subfloat[]{\label{app_3}\includegraphics[width=0.5\textwidth]{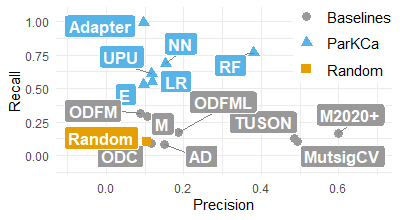}}
    \subfloat[]{\label{app_4}\includegraphics[width=0.5\textwidth]{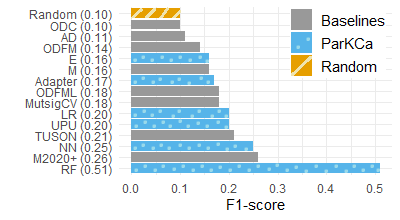}}
        \caption{\textit{(Real-world dataset)} Comparison between driver gene discovery baselines and ParCKa. The goal is to predict driver genes (causal variables) correctly. Large Recall, Precision, and F1-score indicate good models. The score reported is from the full real-world dataset.
    }
    \label{fig4}
\end{figure}

\textbf{Comparison between ParKCa and other baselines}: We compared our results from the real-world dataset with reported results from eight driver gene discovery methods analyzed using the Cancer Gene Census (CGC)\cite{tokheim2016evaluating}. The baselines are MutsigCV, ActiveDriver (AD), MuSiC, OncodriveCLUST (ODC), OncodriveFM (ODFM), OncodriveFML (ODFML), TUSON, and 20/20+. Their approaches vary from analysis of somatic point mutations, mutation significance, functional impact and clusters of somatic mutations, and Random Forest of previous driver genes methods. These baselines are not considered causal discovery methods, which is why we did not use them as learners, but are strong methods that try to solve the same problem on the real-world dataset. We remind the reader that ParKCA takes partial knowledge of causal genes and genomic data as input, while the compared methods only have genomic data as input, but use multiple and more sophisticated types of genomic data. We used the results reported by Tokheim et al\cite{tokheim2016evaluating}.  

It is important to point out that the choice of what is considered a good driver gene discovery model is also an open question. Large recall and small precision indicate models that can recover many known driver genes at the cost of a high rate of False Positives (FP). One can interpret this as a bad model because the true driver genes are lost in the middle of the FP, while others might think that this is an indication of a larger number of unknown driver genes yet to be discovered and explored. On the other hand, high precision and low recall indicate models good at identifying certain driver genes; however, they fail to identify a broader range of them, reflected by a large number of False Negatives (FN). The F1-score summarizes these measures by giving the same importance to both of them. Figure \ref{app_3} indicates that ParKCA meta-learners have a larger recall (0.69 on average) and smaller precision (0.16 on average). On the other hand, the baselines have lower recall (0.17 on average) and larger precision (0.28 on average). Figure \ref{app_4} shows that ParKCA with RF has the largest F1-score, which is almost the double of the largest F1-score from the baseline methods. Overall, ParKCa has competitive results when compared to existing driver gene discovery methods.

\section{Discussion and Conclusion}

Our proposed method ParKCa demonstrated excellent results in the experiments. For small percentages of known causes, Adapter-PU was the meta-learner with the best performance. Furthermore, there was almost no difference between PU models and traditional classification models when the percentage of known causes was above 70\%. If the unlabeled examples are mostly negative, the contribution that PU methods can make is limited, and PU classification reduces to traditional binary classification. 

While our simulations show the efficacy of our method, we highlight that in practice, the results crucially depend on the list of known causes. If this list is comprehensive and includes causes with diverse behaviors, ParKCa will likely succeed in its task. On the other hand, if the list is biased towards certain characteristics, our method might only identify causes with behavior similar to the known examples. Furthermore, ParKCa performance also relies on the assumptions of the level 0 learner's being met. 

In conclusion, we believe our proposed method ParKCa makes important contributions to the causal discovery and causal inference. ParKCa exploits partial knowledge of causes, is flexible, robust, easy to use, and demonstrated promising results on a real-life dataset and in simulations. After narrowing down from thousands or millions of potential causes, the causes detected by ParKCa can be better explored and confirmed thought rigorous laboratory studies. Improvements in causal discovery and causal inference methods that work on CB applications can bring many other benefits beyond the development of sophisticated causal discovery techniques. A successful approach has the potential to significantly improve therapy recommendations, working towards the goal of precision medicine to provide the ``right drug at the right dose to the right patient''\cite{collins2015new}.\\

\noindent\textbf{Supplemental Material:}\\
\url{sites.google.com/view/raquelaoki/publications/parkca-supplemental-material}\\
Code: \url{https://github.com/raquelaoki/ParKCa}

\bibliographystyle{ws-procs11x85}
\bibliography{ws-pro-sample}

\end{document}